\documentclass{article}
\usepackage{ijcai21}

\usepackage{times}
\usepackage{soul}
\usepackage{url}
\usepackage[hidelinks]{hyperref}
\usepackage[utf8]{inputenc}
\usepackage[small]{caption}
\usepackage{graphicx}
\usepackage{amsmath}
\usepackage{amsthm}
\usepackage{booktabs}
\usepackage{acronym}
\usepackage{multirow}
\usepackage{comment}
\usepackage{subcaption}
\usepackage{amssymb}
\usepackage{relsize}
\usepackage{xcolor}
\usepackage{todonotes}

\usepackage[switch]{lineno}
\DeclareMathOperator{\Fr}{Fr}

\DeclareMathOperator{\iou}{IoU}
\DeclareMathOperator{\thhhh}{th}
\DeclareMathOperator{\Ra}{R_{bn}}
\DeclareMathOperator{\Rb}{L_g}
\DeclareMathOperator{\Rc}{L_s}

\DeclareMathOperator{\scr}{\mathcal{D}}
\newcommand\blfootnote[1]{%
  \begingroup
  \renewcommand\thefootnote{}\footnote{#1}%
  \addtocounter{footnote}{-1}%
  \endgroup
}
\newcommand*{\teal}{\textcolor{black}}

\acrodef{cnn}		[\textsc{CNN}]				{convolutional neural network}
\acrodef{cnns}		[\textsc{CNN}s]				{convolutional neural networks}
\acrodef{rf}        [\textsc{RF}]               {Receptive Field}
\acrodef{iou}       [\textsc{IoU}]              {Intersection over Union}

\title{Learning Interpretable Concept Groups in CNNs}

\author{
	Saurabh Varshneya$^1$\footnote{Contact Author: varshneya@cs.uni-kl.de}
	\and
	Antoine Ledent$^1$\and
	Robert A. Vandermeulen$^2$\and
	Yunwen Lei$^3$\and
	Matthias Enders$^4$\and
	Damian Borth$^5$\And
	Marius Kloft$^1$\\
	\affiliations
	$^1$Technical University of Kaiserslautern, Germany\\
	$^2$Technical University of Berlin, Germany\\
	$^3$University of Birmingham, United Kingdom\\
	$^4$NPZ Innovation GmbH, Germany\\
	$^5$University of St.Gallen, Switzerland\\
\emails
\{varshneya, ledent, kloft\}@cs.uni-kl.de,
vandermeulen@tu-berlin.de,
y.lei@bham.ac.uk,
m.enders@npz-innovation.de,
damian.borth@unisg.ch
}

\usepackage{xr}

\begin{document}

	\maketitle

	\begin{abstract}
		We propose a novel training methodology---Concept Group Learning (CGL)---that encourages training of interpretable CNN filters by partitioning filters in each layer into \emph{concept groups}, each of which is trained to learn a single visual concept. We achieve this through a novel regularization strategy that forces filters in the same group to be active in similar image regions for a given layer. We additionally use a regularizer to encourage a sparse weighting of the concept groups in each layer so that a few concept groups can have greater importance than others. We quantitatively evaluate CGL's model interpretability using standard interpretability evaluation techniques and find that our method increases interpretability scores in most cases. Qualitatively we compare the image regions that are most active under filters learned using CGL versus filters learned without CGL and find that CGL activation regions more strongly concentrate around semantically relevant features.\blfootnote{Code will be available at: \url{https://github.com/srb-cv/cgl}}
		
	\end{abstract}

	\section{Introduction}
	
    There is a great interest in understanding the hidden representations produced by \ac{cnns}. Understanding these representations has significant implications theoretically (a better understanding of neural networks in general) and practically (trustworthiness in AI). Toward this end, many methods for interpreting hidden representations have been proposed. We delineate two avenues of research, which include (1) visualizing the regions of images that lead to high activation responses in a filter~\cite{NIPS2014_5347,DBLP:journals/corr/abs-1710-10577,mahendran2016visualizing} and (2) interpreting the semantic meaning learned by a filter by finding its alignment to human-interpretable visual concepts, such as object, scene, and color~\cite{netdissect2017,Zhou2018InterpretingDV,zhang2020interpretable}.  

	There exists much work that focuses on understanding the hidden interpretations of \ac{cnns}, but there is a lack of work that actively influences the learning process to favor interpretable representations. 
	
	To address this gap we propose a framework for CNNs that not only \textit{assesses} interpretability but \textit{induces} it. Our approach is to induce a group structure in \ac{cnns} at training, where groups of filters with a useful semantic meaning emerge during the training process. 
	    \begin{figure}[!b]
		\centering
		\includegraphics[width=0.47\textwidth]{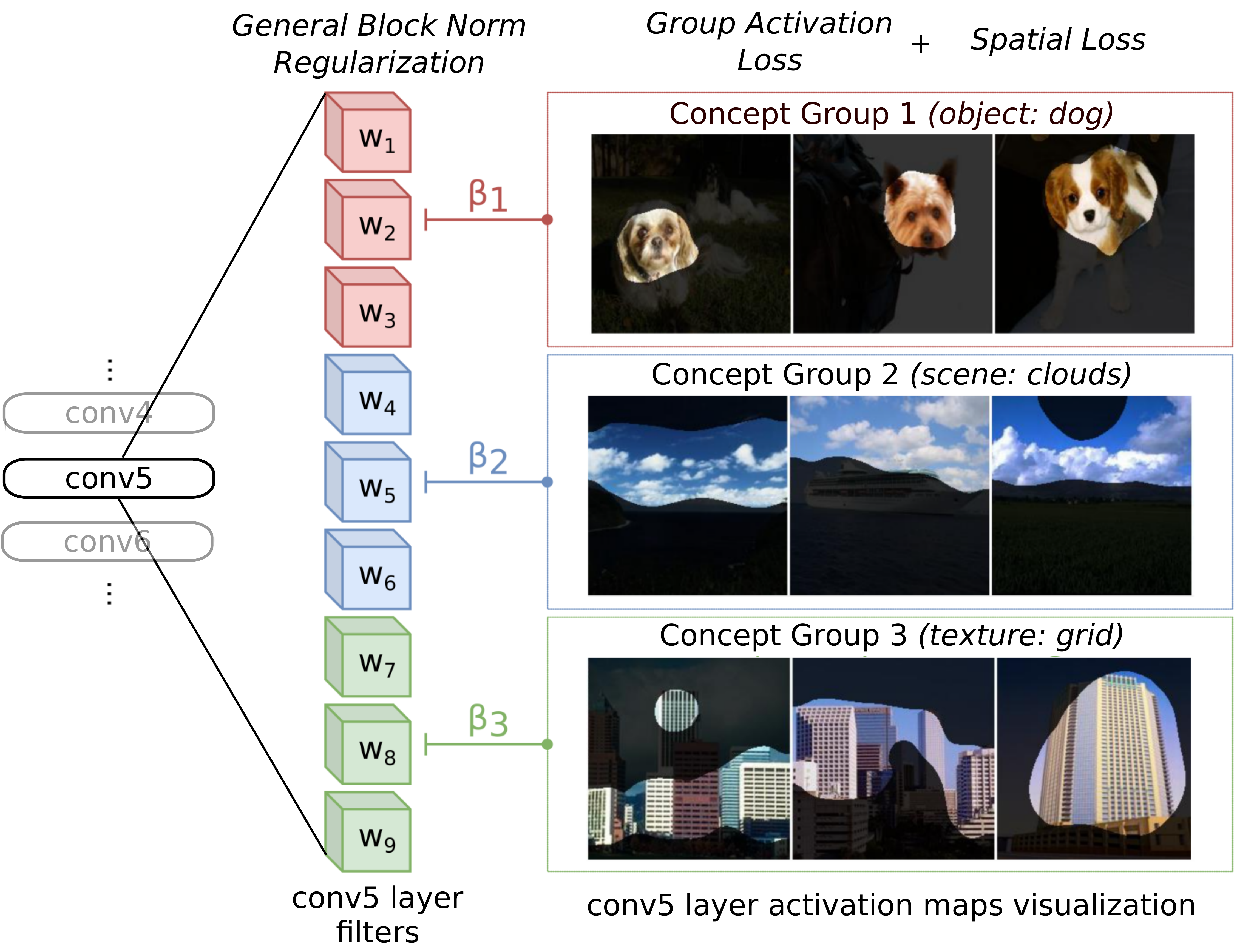}
		
		\caption{An illustration of our framework of learning concept groups in a convolutional layer of a CNN.}
		\label{fig:illustration}
	\end{figure}
    We achieve this through carefully chosen regularization and auxiliary loss functions. 
    In our method, before any training is performed, we partition the collection of filters in each layer into groups, which we term \emph{concept groups}. During training we promote filters within the same concept group to learn similar features, so that, after training, they form a group of correlated filters that jointly encode an abstract visual concept (e.g., an object, scene, or color). 
  
    To learn these concept groups, we introduce three novelties to our training objective (illustrated in Figure~\ref{fig:illustration}):
    1. We propose a new loss function (entitled \emph{group activation loss}) that encourages the filters within each concept group to have highly overlapping activation regions for a given sample. 
    2. We propose another loss function (entitled \emph{spatial loss}), which promotes the activations of each filter to be concentrated around their mean position. 
    3. Using an $\ell_{2,1}$-block-norm regularizer on the concept groups, we learn a sparse weighting of the concept groups. These weights can be interpreted as the relative importance of each concept group.
    \teal{A key property of our method is that it is completely unsupervised, automatically finding concepts using zero side-information. In practice one would likely include our losses in conjunction with some other task, e.g. classification, to increase interpretability of that task, but none of the additional task information, e.g. class labels, is used in our losses.}

	
	 In addition to the training method described above, it can be observed that interpretability of \ac{cnns} decreases when batch-normalization layers are introduced~\cite{Zhou2018InterpretingDV}. To remedy this we incorporate a variation of batch-norm which works together with its regularizer and losses to achieve the desired group structure in the hidden representations and enhance interpretability.
	We perform a quantitative analysis following the experimental setup by~\cite{netdissect2017} and find that our method significantly improves interpretability. We also analyze the filters of our model qualitatively by visualizing the filters' activations and find that our training setup yields representations with noticeably more interpretable filters.

	\section{Related Work}
	We mention here a few existing works that are related to our training method and outline the major differences with the existing approaches.
	\paragraph{Group convolutions and structured sparsity.} Group convolutions were first used in Alexnet~\cite{krizhevsky2017imagenet} by distributing filters into groups to train the model over multiple GPUs. It is important to note that our concept group approach is different from group convolutions, where there is no connection among the activation maps of different partitioned groups. In contrast, concept groups' activation maps are concatenated at each layer before feeding them to the next layer (see Figure~\ref{fig:general_block_norm}).
	
	There exist methods that induce structured sparsity in CNNs~\cite{wen2016learning,li2017pruning}. Such methods show that inducing sparsity among groups of filters can achieve state-of-the-art accuracy on various datasets with reduced computational costs for CNNs. However, less is known about the effect of such a regularization on the interpretability of networks.
	\paragraph{Disentangled representations.} For improving hidden representations while training,  a group of methods endeavours to obtain \emph{disentangled representations}. The aim of such methods coincides with the main focus of our research. These methods aim to learn better representations where each filter represents a clear and unique visual concept. \teal{\cite{Zhang_2018_CVPR} introduce a training method which aims to train each filter in the final convolutional layers of a CNN to represent an object part. The activation loss from their method encourages a filter to activate only for the training images that belong to the assigned category in the training set.}
	
	\teal{Our own techniques also rely on applying a loss to the activation maps of filters to inject priors into the CNN representation. However, there is a major difference in how we apply the losses on activations in our study. 
	In \cite{Zhang_2018_CVPR} each filter is pushed towards a \emph{known} concept corresponding to an object part.
	Ultimately the \emph{concepts} that this method extracts are parts of the objects in the training classes. This is unlike our method which can capture \emph{arbitrary} concepts (not only ones corresponding to object parts) and without any supervision}. Furthermore, the method from \cite{Zhang_2018_CVPR} filters out activations of low magnitude entirely in the forward pass. We argue that such hard constraints on the activation map can hamper the representation power of the network. To retain representation power our method applies soft constraints on the activations. This does not require any changes in the forward pass.
	
	
	The rest of this paper is organized as follows: in Section~\ref{section:algorithm}, we mathematically define our algorithm and precise regularization procedure to make the CNNs more interpretable. In Section~\ref{section:evaluation}, we explain the evaluation techniques by~\cite{netdissect2017}, which we later use to evaluate the interpretability of the networks trained with our own training methods. Finally, in Section~\ref{section:accuracy}, we compare the results of the experiments we performed with the existing state-of-the-art methods.
	
	\section{Algorithm}
	\label{section:algorithm}

\begin{figure}[ht]
		\centering
		\includegraphics[width=0.45\textwidth]{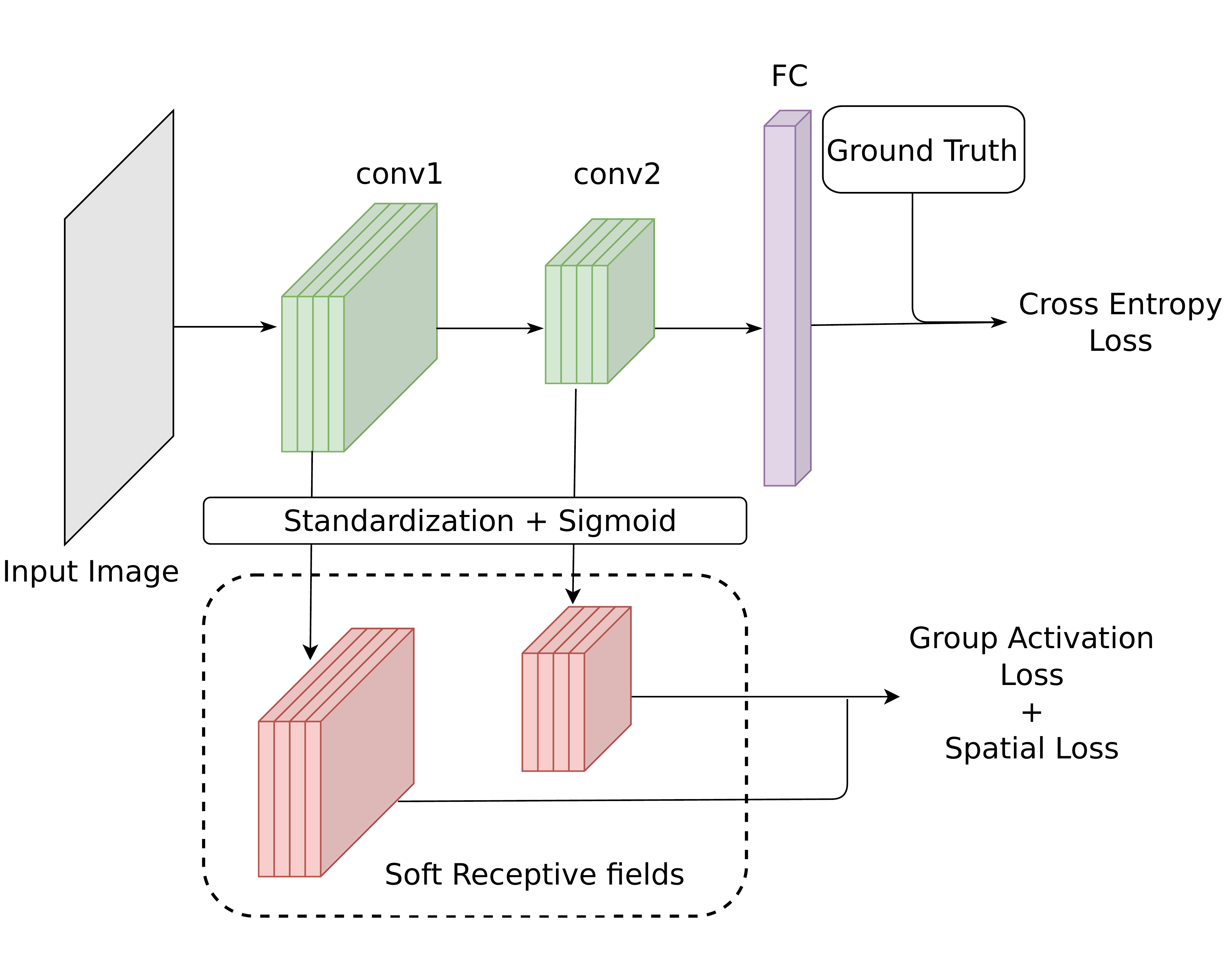}
		\caption{The training setup for our auxiliary loss functions. At every layer we obtain a soft receptive field by feeding a suitably standardized version of the activation maps through a sigmoid. We apply constraints on the soft receptive fields by our well-defined losses}
		\label{fig:loss_functions}
	\end{figure}

    Here we present our method which induces interpretable filters in all layers of a CNN. We briefly outline our method here and describe them precisely in the sequel. Central to our method is an initial grouping of each convolutional layer's filters into $G$ equal-size groups, e.g. for a collection of filters in a layer indexed by $1,\ldots, F$ we divide the indices into $G$ groups $(1,\ldots,f),(f+1,\ldots, 2f),\ldots ((G-1)f+1,\ldots, Gf)$ with $F=Gf$. During training we aim to have a single group of filters (e.g. $(f+1,\ldots, 2f)$) correspond to a single concept. We assign multiple filters to the same concept in the hopes of robustly capturing the concept in a way that would not be possible with a single filter. The number of groups is a hyperparameter that is determined before training. To make the method less restrictive, we can leave some ``free" filters outside the grouping. We consider each concept as a block matrix and apply a general block norm over the collection of groups so as to encourage structured sparsity~\cite{wen2016learning}. The filters belonging to each group are trained to correspond to similar visual concepts. To achieve this we feed each filter's activation map through a sigmoid to determine a (soft) receptive field which represents the areas of the image where the filter is active and then apply a loss to induce filters in the same concept group to have receptive fields that largely overlap and are concentrated to a connected region within an image (illustrated in Figure~\ref{fig:loss_functions}). We now precisely describe the regularizers in the following.

	\subsection{Group Activation Loss \texorpdfstring{$\Rb$}{Lg}}
	\label{sec:group_activation_reg}
	Here we describe our loss function that encourages filters in the same concept group to learn similar concepts. To do this our loss function, which we denote by $\Rb$, will encourage filters within the same concept group to be active in roughly the same area for each input image. 
 	To determine whether a filter is ``active'' in a region of the image, we first feed the activation maps before applying any non-linearity (we call this the \emph{pre-activation map}) through a linear scaling (which is also learned and shared between layers) followed by a sigmoid. We interpret the output of the sigmoid as the probability of a filter being active at a location.

	Let us denote the activation map of the $i^{\thhhh}$ filter in group $g$ in layer $l$ by
	$a_{ig}^l \in \mathbb{R}^{A_l}.$ Here, $A_l$ denotes the size of the pre-activation maps in layer $l$. A filter and its activation maps are typically tensors but we simply consider a flattened version of them for notational simplicity. 


    From our pre-activation map we utilize a sigmoid to yield a \emph{soft-receptive field}, $\psi_{ig}^l \in \left(0,1\right)^{A_l}$, which is defined by the following:
	\begin{equation}
		\psi_{ig}^l =\sigma \left( P_1  \frac{a_{ig}^l}{ S_{ig}^l} + P_2 \right),
	\end{equation}
	where $S_{ig}^l$ is the standard deviation of the  $i^{\thhhh}$ activation-map of group $g$, computed over the minibatch. Note that $P_1$ and $P_2$ are a pair of scaling parameters which can be learned during training. These parameters are shared for all the activation maps in the network. We now define the distance between  two soft-receptive fields $\psi_1$ and $\psi_2$ as 
	\begin{equation}
		\scr(\psi_{1}, \psi_{2}) = \frac{2\|\psi_{1} - \psi_{2}\|_{1}}{\|\psi_{1} \|_{1}+\| \psi_{2}\|_{1}+\|\psi_{2} - \psi_{1}\|_{1}},
	\end{equation}
	where $\|.\|_1$ denotes the $L^1$ norm: 
	$\|x\|_1=\sum_{i}|x_i|$.
	
	The motivation of this distance is to create a soft equivalent to one minus the $\iou$ (intersection over union). This score is frequently used in computer vision to assess the performance of tracking or localization algorithms~\cite{Tychsen-Smith_2018_CVPR,huang2019batching}. To see the equivalence, suppose that $\psi_1$ and $\psi_2$ are the indicator functions of two sets $A$ and $B$. Then we have 
	\begin{align}
	   &\mathcal{D}(\psi_1,\psi_2)=\frac{2\|\psi_1-\psi_2\|_1}{\|\psi_1\|_1+\|\psi_2\|_1+\|\psi_1-\psi_2\|_1}\nonumber \\&=\frac{2\mu(A\Delta B)}{\mu(A)+\mu(B)+\mu(A\Delta B)} 
	   =\frac{2\mu(A\Delta B)}{2\mu(A\cup B)}\nonumber \\
	   &=1-\frac{\mu(A\cap B)}{\mu(A\cup B)}=1-\text{IOU},
	\end{align}
	where we use the standard notation $A\Delta B:= (A\setminus B) \cup (B\setminus A)=(A\cup B )\setminus(A\cap B)$ for the symmetric difference between $A$ and $B$, and $\mu(.)$ denotes the Lebesgue measure or the cardinality in the case of finite sets. 
The definition of $\psi$ must be altered slightly to work with batch normalizaiton; we describe this in the following.
	
	\subsubsection{Definition of \texorpdfstring{$\psi$}{psi} With Batch-normalization}
	If the network is trained with batch-norm, the pre-activations are standardized to have mean $0$ and variance $1$. The input to the next layer is then $ \gamma_{ig}^l(a_{ig}^l)+\beta_{ig}^l$, where $\beta_{ig}^l$ and $\gamma_{ig}^l$ are the learned parameters of a batch-norm layer over the $i^{\thhhh}$ activation-map, of group $g$, as in standard batch normalization~\cite{Ioffe:2015:BNA:3045118.3045167}. To modify the batch normalized values to be compatible with our method we first define a variable $\tau_{ig}^l$ by
	$$
	\tau_{ig}^l = a_{ig}^l + \frac{(\beta_{ig}^l)}{(\gamma_{ig}^l)},
	$$ 
	which can be thought of as undoing batch normalization for the purposes of finding a soft receptive field, since batch normalization is known to reduce interpretability~\cite{netdissect2017}. We can now feed $\tau_{ig}^l$ through a sigmoid to obtain our soft receptive field as follows:
	\begin{equation}
	    \psi_{ig}^l = \sigma\left(P_1  \tau_{ig}^l+ P_2\right).
	\end{equation}

	\subsubsection{Formula for \texorpdfstring{$\Rb$}{Lg}}
	\label{R1}
	We can now define our group activation loss $\Rb$ to be the following (note that the second term is new and is described in the next paragraph)
	\begin{align}
    	&\frac{1}{r}
    	\dfrac{\sum_{l,g}  \sum_{(i,j)\in \mathcal{R}} \;\;\;\;\;\;\;\;\;\;\;\;\; 2\|\psi_{{j}g}^{{l}} - \psi_{ig}^{l}\|_{1}\;\;\;\;\;\;\;\;\;\;\;\;\;\;\;}{\sum_{l,g}\sum_{(i,j)\in \mathcal{R}}   \|\psi_{{j}g}^{{l}} \|_{1}+\| \psi_{ig}^{l}\|_{1}+\|\psi_{{j}g}^{{l}} - \psi_{ig}^{l}\|_{1}}+ \cdots \nonumber
    	\\
    	&
    	\lambda_{2}
    	\frac{1}{r}
    	\dfrac{\sum_{l,g}\sum_{(i,j)\in \mathcal{R}} \;\;\;\;\;\;\;\;\;\;\;\;\;2\|\psi_{{j}g}^{{l}} - \psi_{ig}^{l+1}\|_{1}\;\;\;\;\;\;\;\;\;\;\;\;\;\;\;}{\sum_{l,g}\sum_{(i,j)\in \mathcal{R}}\|\psi_{{j}g}^{{l}} \|_{1}+\| \psi_{ig}^{l+1}\|_{1}+\|\psi_{{j}g}^{{l}} - \psi_{ig}^{l+1}\|_{1}} \label{eqn:rb}
	\end{align}
	where $\mathcal{R}$ is a random subset of cardinality $r$ of $\{1,2,\ldots, N_g^{l}\}\times \{1,2,\ldots, N_g^{l}\}$ that changes at each iteration of the gradient descent procedure. $N^{l}_g$ denotes the number of activation maps in group $g$ in layer $l$. We use this randomization over $\mathcal{R}$ as an approximation of comparing all pairs of filters in each concept group which is computationally expensive. Here we write $\psi_{jg}^l$ for the vector composed of the components $\psi_{jg}^l(x_i)$ for samples $x_i$ in some minibatch. All the $L^1$ norms are computed over the minibatch  (as well as the spatial dimension).  The value for $r$ can be set as a hyperparameter. We find that setting $r$ to $3N_g^{l}$ works well in practice. 
	
	
	Each term in the summation that appears in $\Rb$ is a soft version of one minus the intersection over union measure of similarity between the receptive fields given by $\scr(\psi_{ig}^l, \psi_{jg}^l)$.
	As some concepts typically gradually change or die out in higher layers, we additionally consider employing a sliding comparison, the second term in \eqref{eqn:rb}, where the filters in group $g$ of layer $l$ are required to be similar to those in group $g$ of layer $l+1$ but are not directly compared to those in group $g$ in layers beyond $l+1$. We found that enforcing this similarity of concepts throughout layers is too strong a requirement and simply causes the filters to go to zero. Because of this we use $\lambda_{2}=0$ for all $l$ in our experiments.

	
	\subsection{Spatial Loss \texorpdfstring{$\Rc$}{Ls}}

	
	In this section we introduce a loss that aims to regularize each filter to activate on a single, connected area of each image. Suppose for instance that an image contains a car and a house both of which are relevant concepts to the task to which the CNN is being applied. Since both concepts are abstract they are likely to be represented near the final convolutional layers of a CNN where the receptive fields are large. We would like to keep these concepts separate between concept groups and prevent them from incorporating large portions of the image corresponding to multiple abstract concepts. To achieve this we enforce each filter to only be active on a small connected region on images by penalizing activation responses that far from the center of activation. The center of activation is computed over the soft receptive field $\psi$ for each activation map. We define spatial loss $\Rc(\psi)$ as the following:
 	\begin{equation*}
 	    \Rc(\psi)=\frac{\sum_{j} \psi_{j}\|j-c(\psi)\|_2}{\sum_{j} \psi_{j}},\ \text{where }c(\psi)=\frac{\sum_{j}j\cdot\psi_{j}}{\sum_{j}\psi_{j}}
 	\end{equation*}
 	and $j$ denotes the index of a activation of the soft receptive field ($\psi$) of a filter. Here $c$ represents the mean position of activation of a soft receptive field and $\Rc$ is analogous to the corresponding variance. The loss is computed over all the activation maps of a filter obtained in a minibatch.


		\begin{figure}[hb]
		\centering
		\includegraphics[width=0.48\textwidth]{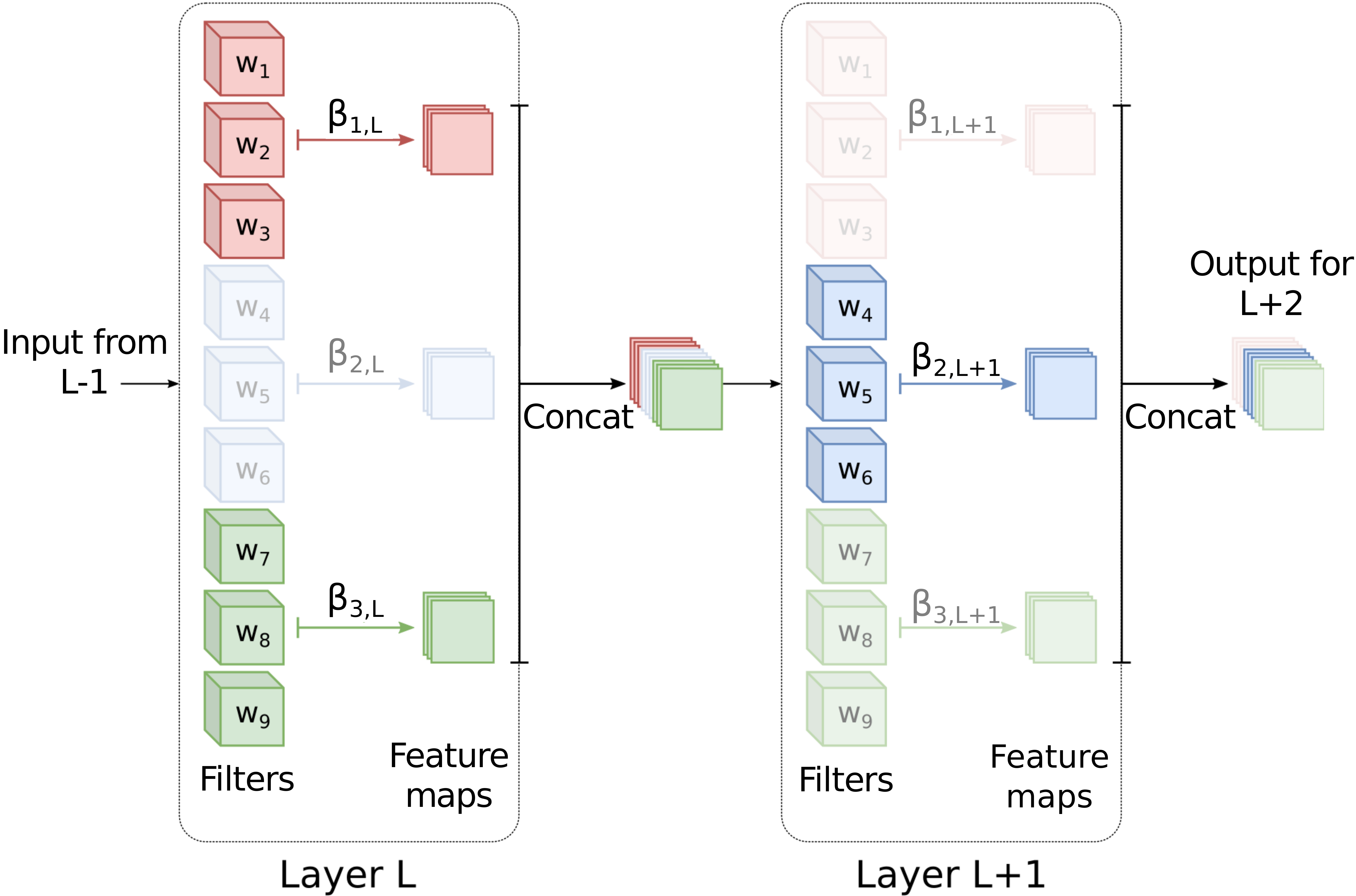}
		\caption{The general block norm induces a group structure in the filters of a layer, where few groups are dominantly learned over other groups. $\beta_1$, $\beta_2$ and $\beta_3$ are the relevance factors of each group which are learned while training a \ac{cnn}}
		\label{fig:general_block_norm}
	\end{figure}
	
	\subsection{General Block Norm Regularizer  \texorpdfstring{$\Ra$}{R1}}
	
	In this section we define our general block norm regularizer $\Ra$ that replaces the conventional weight decay regularizer. This changes the inductive bias of our training algorithm so that it induces a group sparse structure over the concept groups. This regularizer encourages some concept groups to be of greater importance than others, thereby giving a notion of importance to concept groups and encouraging our concept groups to be parsimonious. This allows us to avoid situations where, for example, multiple concept groups capture the concept of ``car.'' Such sparse models also yield reduced memory requirements and improved training speed as side benefits~\cite{mocanu2018scalable}. An illustration the result of using this regularization is shown in Figure \ref{fig:general_block_norm}. We will denote by $w_{gl}$ the matrix containing all the weights of the filters of group $g$ in layer $l$.
	The regularizer $\Ra$ can then be written as :
	\begin{equation}
		\Ra = 
		\sum_{l=1}^{L}
		\sum_{g=1}^G
	\|w_{gl}\|_{\Fr},
	\end{equation}
	where $L$ denotes the number of convolutional layers in the network, $G$ the number of concept groups per convolutional layer, and $\|\cdot\|_{\Fr}$ denotes the Frobenius norm i.e., $\|M\|_{\Fr}^2=\sum_{i,j} M_{i,j}^2$ for a matrix $M$.

	An advantage of using the general block norm regularizer $\Ra$ is that we can obtain a vector of relevance factors $\beta$ for our concept groups, as shown in Figure~\ref{fig:general_block_norm}. These relevance scores can be obtained by a summation of the norms as follows: 
	\begin{equation}
		\label{eq:relevance_weight}
		\beta_{gl} = \|w_{gl}\|_{\Fr}
		\quad \text{and} \quad \beta_l=\sum_{g=1}^{G}\beta_{gl},
	\end{equation}
	where, $\beta_{gl}$ denotes a group's relevance in layer $l$, and  $\beta_l$ denotes the relevance of layer $l$ in a network.
	\paragraph{Complete framework.}
	In summary our overall framework is obtained by combining the above mentioned regularization strategy and the two loss functions. For a simplified view of the combined method, we consider a \ac{cnn} with weight parameters $w$ and we denote the soft receptive fields obtained from the activation maps by $\psi$. The final optimization target for an interpretable \ac{cnn} can be defined as:
	\begin{equation}
		\mathbf{L} = L_{d}(w) + \lambda_{bn} \Ra(w) + \lambda_g \Rb(\psi) + \lambda_s \Rc(\psi),
	\end{equation}
	where, $L_d$ denotes the main loss for our task (e.g. cross entropy for classification), $\Ra$ is the general block norm regularization, $\Rb$ and $\Rc$ are the group activation loss and spatial loss respectively. $\lambda_{bn}$, $\lambda_g$ and $\lambda_s$ are the weightings of the defined regularization and auxiliary losses with respect to the main loss. 
	
	
	\section{Experiments}
	In this section we explain the evaluation strategy we adopt for the quantitative and qualitative analysis of the interpretability of CNNs, and describe the results of our experiments performed on synthetic and real world datasets.

	\subsection{Interpretability: Quantitative Assessment} 
	\label{section:evaluation}
	In order to quantify the interpretability of a trained \ac{cnn}, we exactly replicate the evaluation on the Broden dataset as proposed by~\cite{netdissect2017}. The Broden dataset contains pixel-wise annotations of a broad range of categories, which belong to one of six human-interpretable visual concepts. We compute the alignment of a filter with a category by comparing its activation maps for a set of images against the available ground truth using the same threshold and scoring function described by~\cite{netdissect2017}. The interpretability of a layer can then be defined as the number of unique categories assigned to the filters in that layer. Authors call this as the number of \textit{unique detectors}.

	\subsubsection{Evaluation on a Synthetic Dataset}
	
	In order to validate the efficacy of our methods in a controlled environment, we constructed a synthetic dataset with obvious human-interpretable visual concepts corresponding to shape and color: each image contains two randomly positioned figures among the set \{square, circle, triangle\} with repetitions allowed, and each shape is assigned a colour from the set \{red, blue, green\}. Note that in principle, there are 15 high-level concepts in this paradigm: one for each combination of colour and shape, and one for each colour and each shape separately. 
	
	We consider two training settings: in the first case, we have one label for each combination of shapes and colours (there are therefore 45 labels). In the second setting, we simply assign the label one if a square is present and zero otherwise (the problem is therefore binary). Therefore, in the first setting, all the information contained in the high level concepts ``shape, color, color-shape" is contained in the label, whereas in the second situation, only part of this information is encoded in the label. 
	
	\teal{For training, we use a simple CNN with $2$ convolutional layers, having $128$ and $256$ filters in the first and second layer respectively.
    We set the size of filters to $3 \times 3$ for both the layers.
    }The evaluation is performed using the same metrics as described above. We report the results of the second, binary setting in Table~\ref{tab:synthetic1}. In the multiclass setting, the network is able to recover all concepts very well regardless of regularization. However, we find that in the binary setting, the addition of our regularizers has a very strong positive effect on the recovery of the concepts: in the case where we add all three regularizers, 14 out of 15 of all unique concepts are recovered at the second layer, compared to only 10 when using weight decay. 
		
	This is a significant illustration of the fact that our method is able to guide the network towards a more complete and disentangled representation of all salient features of the data in the absence of low-level feature annotations.

	\begin{table}[ht]

		\centering
		\footnotesize\addtolength{\tabcolsep}{-1pt}
		\begin{tabular}{c c r r r r}
			\toprule
			{Regularizer} &  {layer} & \multicolumn{4}{c}{{No. of Unique Detectors}}\\
			\midrule
			{}            & {}   & {color} & {shape} & {color-shape} & {Total}\\
			\multirow{2}{*}{Weight-Decay}     &  conv1      & 3   &  0 & 4 & 7 \\
			{}                               &  conv2  & 3   &  0 &7&10\\
			\hline
			\multirow{2}{*}{$\Ra$}  &  conv1      &    3 & 2& 4&9\\
			{} &  conv2 &  3& 2&7&12\\
			\hline
			\multirow{2}{*}{$\Ra + \Rb + \Rc$}  &  conv1      & 3 & 3& 5&11\\
			&  conv2  &  3& 2&9&14\\
			\bottomrule
		\end{tabular}
		\caption{Number of unique detectors in each layer while training on synthetic data in a two class labels setup}
		\label{tab:synthetic1}
	\end{table}

	\subsubsection{Evaluation on the Real Datasets}
    We tested our methods on three well-known convolutional architectures named Alexnet, Alexnet-B and VGG, where Alexnet-B is the version containing the batch normalization layers added to the common Alexnet. All the networks are trained from scratch on the two well-known image classification datasets Places365~\cite{zhou2017places} and ImageNet~\cite{deng2009imagenet}. The Places365 dataset is comparable to the ImageNet dataset, which contains 1.8 million training images with 365 scene categories. For evaluation, we exactly replicate the method from~\cite{netdissect2017}, as explained above to compute the interpretability of the trained models. For a baseline of interpretability, we obtained the pretrained models for Imagenet and Places365 datasets from the Pytorch and MIT CSAIL repositories respectively.
    Table~\ref{tab:detectors} shows the comparison of the interpretabilty scores for the models trained with the conventional weight decay regularizer and our training method.
	
	\begin{table}[!b]
		\begin{center}
			\footnotesize\addtolength{\tabcolsep}{-5pt}
			\scalebox{0.85}{
				\begin{tabular}{c r r r r r r r}
					\toprule
					Dataset &  Model & {\footnotesize conv1} & {\footnotesize conv2} & {\footnotesize conv3} & {\footnotesize conv4} & {\footnotesize conv5} & {\footnotesize score}\\
					\midrule
					\rule{0pt}{4ex}
					\multirow{3}{*}{\footnotesize Places365} &     {\footnotesize Alexnet}  & 6 & 12 & 22 & 36 & 54 & 130\\
					{}                         &  {\scriptsize Alexnet, $\mathsmaller {(\Ra)}$ }  & 6 & 16 & 36 & 46 & 69 & 173 \\
					{}                         &  {\scriptsize Alexnet, $\mathsmaller {(\Ra\Rb\Rc)}$ }  & 8 & 17 & 35 & 50 & 75 & 185 \\
					\midrule
					\rule{0pt}{2ex}
					\multirow{3}{*}{\footnotesize Places365}                         &  {Alexnet-B}  & 6 & 13 & 27 & 43 & 59 & 148 \\  
					{}                         &  {\scriptsize Alexnet-B, $\mathsmaller {(\Ra)}$} & 6 & 15 & 30 & 44& 62 & 157\\ 
					{}                         &  {\scriptsize Alexnet-B, $\mathsmaller {(\Ra\Rb\Rc)}$} & 5 & 17 & 30 & 51 & 63 & 166\\ 
					\midrule
					\rule{0pt}{3ex}
					\multirow{3}{*}{\footnotesize Imagenet}                         &  {Alexnet}  & 5 & 20 & 34 & 28 & 49 & 136 \\  
					{}                         &  {\scriptsize Alexnet, $\mathsmaller {(\Ra)}$} & 5 & 16 & 39 & 32 & 45 & 137\\ 
					{}                         &  {\scriptsize Alexnet, $\mathsmaller {(\Ra\Rb\Rc)}$} & 7 & 16 & 34 & 36 & 48 & 141\\
					\midrule 
					\textbf{} &  \textbf{} & {\footnotesize conv3-3} & {\footnotesize conv4-3} & {\footnotesize conv5-1} & {\footnotesize conv5-2} & {\footnotesize conv5-3} & {}\\
					\midrule
					\rule{0pt}{2ex}
					\multirow{2}{*}{\footnotesize Imagenet}  &     {VGG16}   & 10 &  48  & 59 &  53 & 89 & 259\\
					{}                         &  {\scriptsize VGG16, $\mathsmaller {(\Ra\Rb\Rc)}$}  & 14 &  48  & 59 &  62 &  83  & 266\\ 
					\rule{0pt}{3ex}
					\multirow{2}{*}{\footnotesize Places365}  &     {VGG16}   & 9 &  50  & 62 & 75 & 119 & 315\\
					{}                         &  {\scriptsize VGG16, $\mathsmaller {(\Ra\Rb\Rc)}$} & 10 & 45 & 62   & 86  &  129 & 332 \\ 
					\bottomrule
				\end{tabular}
			}
		\end{center}
	\caption{The table above shows the sum of the number of unique detectors in each layer of the network. We achieve a better interpretability across all the tested networks and datasets.}
	\label{tab:detectors}
	\end{table}

	\subsubsection{Analysis over Concept-groups}	
\begin{figure}
		\centering
		\includegraphics[width=0.46\textwidth]{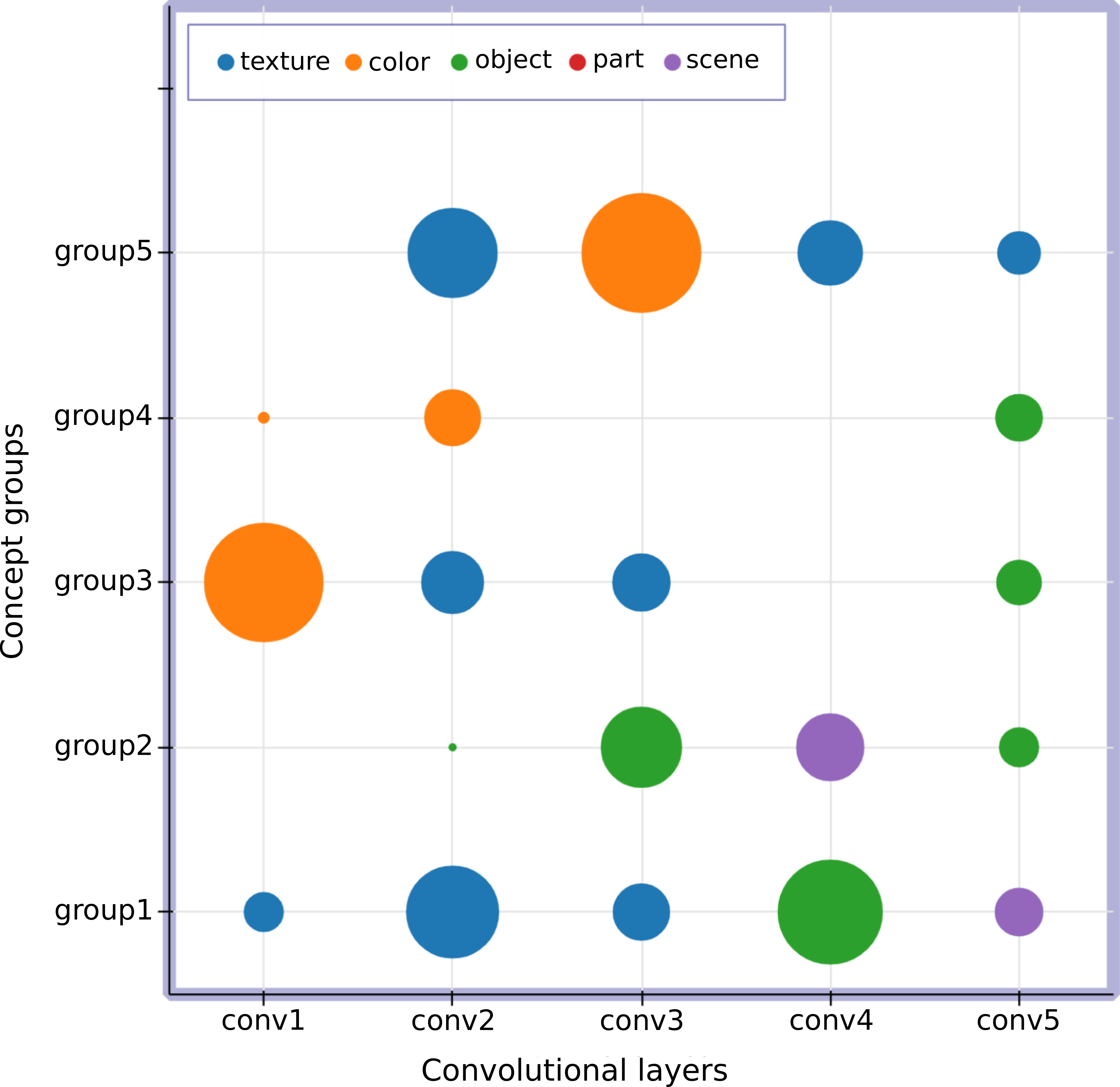}
		\caption{Concept groups that emerge in different layers in a trained Alexnet. The size of the circle represents the relevance of a concept group in a layer.}
		\label{fig:weight_analysis}
	\end{figure}

	For each predefined group, we compute a weighted sum over the number of detectors found in the group and the $\iou$ score. A group is said to be aligned to a visual concept when the weighted sum is greater than a threshold $T$, which was chosen by hand and is the same for all experiments. One can obtain the relevance of such a concept-group from Equation \ref{eq:relevance_weight}. In Figure \ref{fig:weight_analysis}, we show an analysis over the emergent concept-groups in the convolutional layers of Alexnet.

	\subsection{Interpretability: Qualitative Assessment} 
\begin{figure}[ht]
		\centering
		\includegraphics[width=0.44\textwidth]{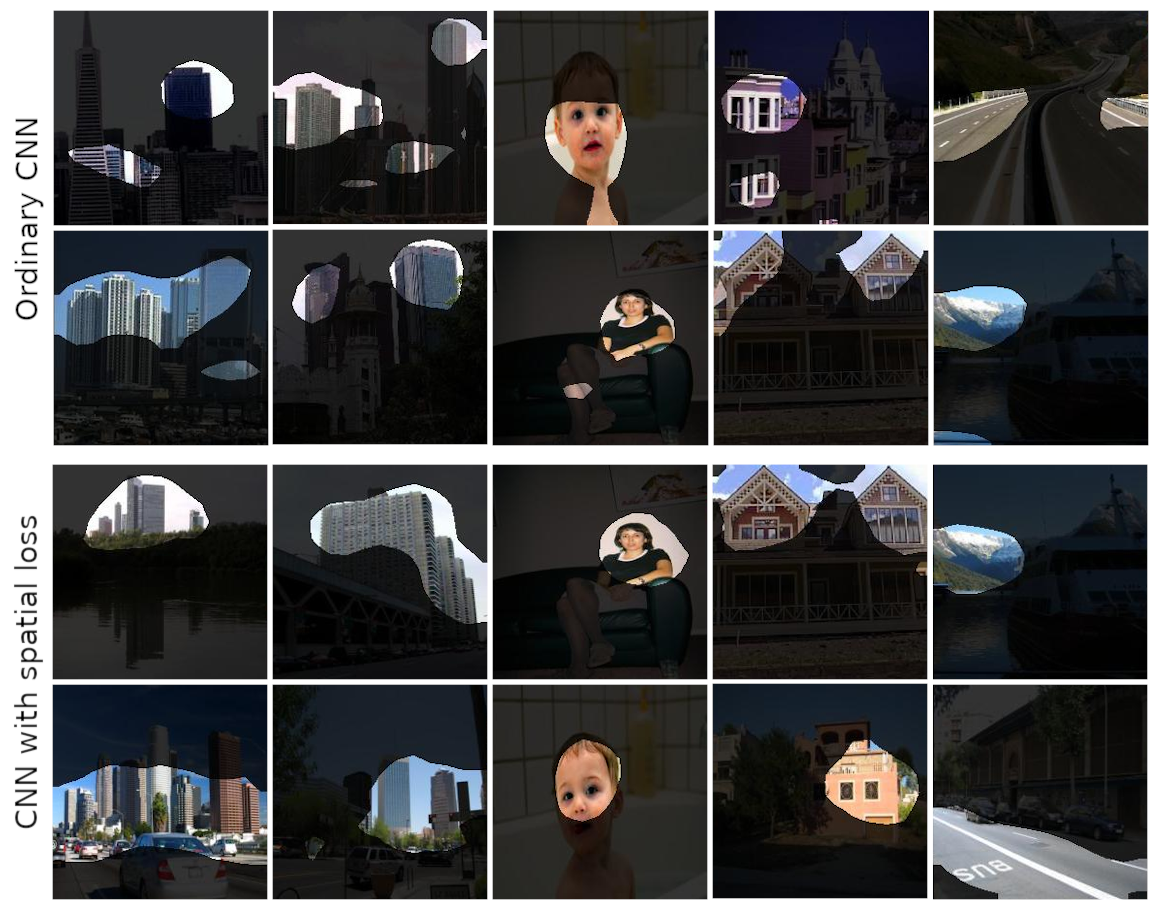}
		\caption{The receptive fields of the filters in the last layer of Alexnet trained with weight decay regularizer (in the top two rows) on Places365 dataset compared with the model trained with spatial loss $\Rc$ (in the bottom two rows).}
		\label{fig:spatial_regularitation}
	\end{figure}

	We use the method described in~\cite{DBLP:journals/corr/ZhouKLOT14} for visualizing filters of trained \ac{cnn}s.  To visualize the receptive field of a filter $f$, its activation maps is computed over a set of images. The maximum activation (max-activation) value for each activation map over all the pixels is extracted. The top $K$ images are then selected corresponding to the activation maps with $K$ largest max-activation values. For each image, regions with ``high" unit activations are identified using a threshold $T_f$---taken from \cite{netdissect2017}. This activated region is then scaled up to the image resolution and the filter $f$'s receptive field can be visualized over the selected images. Overall, this qualitative analysis gives a better understanding of a filter by focusing on important areas in the image~\cite{DBLP:journals/corr/ZhouKLOT14}. We compare the visualizations of the filters of a CNN trained with weight decay against a CNN trained with our proposed losses in Figure \ref{fig:spatial_regularitation}.
	
	\subsection{Accuracy Versus Interpretability}
	\label{section:accuracy}
	Most often, attempts to increase interpretability reduce the discrimination power of a \ac{cnn}, thereby compromising the validation accuracy of the trained network. We show that by applying soft constraints as proposed in this paper, the validation accuracy is only slightly reduced (less than 1 percent in each case). Results in Table~\ref{tab:detectors} show that most of the unique detectors emerge on the last two convolutional layers of every network. Therefore we compare the ratio of unique detectors (RUD), the number of the unique detectors divided by the  total number of filters in a layer, for the final two convolutional layers along with the accuracy of the trained network in Table~\ref{tab:accuracy}.
	
	\begin{table}[ht]
		\centering
		
		\footnotesize\addtolength{\tabcolsep}{-1.5pt}
		\begin{tabular}{l r r r r}
			\toprule
			{Dataset} &  {Network} & {Top-1 acc.} & {Top-5 acc.} & {RUD}\\
			\midrule
			\multirow{6}{*}{\small Places365} &     {\small Alexnet}  & 51.14\% & 81.59\%  & 0.176\\
			{}                         &  {\scriptsize Alexnet, $\mathsmaller {(\Ra \Rb \Rc)}$ }  & 50.46\% & 80.60\% & 0.244\\
			\rule{0pt}{3ex}
			{}                         &  {Alexnet-B*}  & 51.32\% & 81.60\% & 0.199\\  
			{}                         &  {\scriptsize Alexnet-B, $\mathsmaller {(\Ra \Rb \Rc)}$} & 50.46\% & 80.62\% & 0.223\\ 
			\rule{0pt}{3ex}
			{}                         &     {VGG16}   & 54.91\% & 85.02\% & 0.189\\
			{}                         &  {\scriptsize VGG16, $\mathsmaller {(\Ra \Rb \Rc)}$}  & 54.91\% &  85.20\% & 0.210\\ 
			
			\midrule
			\rule{0pt}{2ex}
			\multirow{4}{*}{\small Imagenet}  &     {\small Alexnet}   & 56.30\%  & 79.04\% & 0.150\\
			{}                         &  {\scriptsize Alexnet, $\mathsmaller {(\Ra \Rb \Rc)}$}  & 55.09\% & 77.93\% & 0.164\\ 
			\rule{0pt}{3ex} 
			{}                         &     {VGG16}   & 71.79\% & 90.45\% & 0.139\\
			{}                         &  {\scriptsize VGG16, $\mathsmaller {(\Ra \Rb \Rc)}$} & 70.55\% & 89.87\% &0.142\\ 
			\bottomrule
		\end{tabular}
	\caption{A comparison of the Validation Accuracy and RUD Score on training with different regularizers. The RUD score is the ratio of the total number of unique detectors divided by the number of filters in the last two convolutional layers of a CNN.}
	\label{tab:accuracy}
	\end{table}

	\section{Conclusion}
	Our formulated losses and regularizer exploit the hidden structures of the data via inducing a more prominent group structure among the filters of \ac{cnn}. Even simply replacing the conventional weight decay regularizer by our block norm $\Ra$, we can increase the interpretability of a network without compromising with its discrimination power. We efficiently compute the activation region of a filter over an image in the forward pass as described in Section~\ref{sec:group_activation_reg}, and then constrain the activation regions so that filters in the same group activate similar areas. This further induces a group structure in all the layers of a CNN and enhances interpretability.
	
	\section*{Acknowledgements}
	The authors gratefully acknowledge support by the Carl-Zeiss Foundation, by the German Research Foundation (DFG) award KL 2698/2-1, and by the Federal Ministry of Science and Education (BMBF) awards 01IS18051A and 031B0770E.

	\bibliographystyle{named}
	\bibliography{literature}

\begin{thebibliography}{}

\bibitem[\protect\citeauthoryear{Bau \bgroup \em et al.\egroup
  }{2017}]{netdissect2017}
David Bau, Bolei Zhou, Aditya Khosla, Aude Oliva, and Antonio Torralba.
\newblock Network dissection: Quantifying interpretability of deep visual
  representations.
\newblock In {\em Computer Vision and Pattern Recognition}, 2017.

\bibitem[\protect\citeauthoryear{Deng \bgroup \em et al.\egroup
  }{2009}]{deng2009imagenet}
Jia Deng, Wei Dong, Richard Socher, Li-Jia Li, Kai Li, and Li~Fei-Fei.
\newblock Imagenet: A large-scale hierarchical image database.
\newblock In {\em 2009 IEEE conference on computer vision and pattern
  recognition}, pages 248--255. Ieee, 2009.

\bibitem[\protect\citeauthoryear{Huang \bgroup \em et al.\egroup
  }{2019}]{huang2019batching}
Yifeng Huang, Zhirong Tang, Dan Chen, Kaixiong Su, and Chengbin Chen.
\newblock Batching soft iou for training semantic segmentation networks.
\newblock {\em IEEE Signal Processing Letters}, 27:66--70, 2019.

\bibitem[\protect\citeauthoryear{Ioffe and
  Szegedy}{2015}]{Ioffe:2015:BNA:3045118.3045167}
Sergey Ioffe and Christian Szegedy.
\newblock Batch normalization: Accelerating deep network training by reducing
  internal covariate shift.
\newblock In {\em Proceedings of the 32Nd International Conference on
  International Conference on Machine Learning - Volume 37}, ICML'15, pages
  448--456. JMLR.org, 2015.

\bibitem[\protect\citeauthoryear{Krizhevsky \bgroup \em et al.\egroup
  }{2017}]{krizhevsky2017imagenet}
Alex Krizhevsky, Ilya Sutskever, and Geoffrey~E Hinton.
\newblock Imagenet classification with deep convolutional neural networks.
\newblock {\em Communications of the ACM}, 60(6):84--90, 2017.

\bibitem[\protect\citeauthoryear{Li \bgroup \em et al.\egroup
  }{2017}]{li2017pruning}
Hao Li, Asim Kadav, Igor Durdanovic, Hanan Samet, and Hans~Peter Graf.
\newblock Pruning filters for efficient convnets, 2017.

\bibitem[\protect\citeauthoryear{Mahendran and
  Vedaldi}{2016}]{mahendran2016visualizing}
Aravindh Mahendran and Andrea Vedaldi.
\newblock Visualizing deep convolutional neural networks using natural
  pre-images.
\newblock {\em International Journal of Computer Vision}, 120(3):233--255,
  2016.

\bibitem[\protect\citeauthoryear{Mocanu \bgroup \em et al.\egroup
  }{2018}]{mocanu2018scalable}
Decebal~Constantin Mocanu, Elena Mocanu, Peter Stone, Phuong~H Nguyen,
  Madeleine Gibescu, and Antonio Liotta.
\newblock Scalable training of artificial neural networks with adaptive sparse
  connectivity inspired by network science.
\newblock {\em Nature communications}, 9(1):1--12, 2018.

\bibitem[\protect\citeauthoryear{Tychsen-Smith and
  Petersson}{2018}]{Tychsen-Smith_2018_CVPR}
Lachlan Tychsen-Smith and Lars Petersson.
\newblock Improving object localization with fitness nms and bounded iou loss.
\newblock In {\em Proceedings of the IEEE Conference on Computer Vision and
  Pattern Recognition (CVPR)}, June 2018.

\bibitem[\protect\citeauthoryear{Wen \bgroup \em et al.\egroup
  }{2016}]{wen2016learning}
Wei Wen, Chunpeng Wu, Yandan Wang, Yiran Chen, and Hai Li.
\newblock Learning structured sparsity in deep neural networks.
\newblock {\em Advances in neural information processing systems},
  29:2074--2082, 2016.

\bibitem[\protect\citeauthoryear{Yosinski \bgroup \em et al.\egroup
  }{2014}]{NIPS2014_5347}
Jason Yosinski, Jeff Clune, Yoshua Bengio, and Hod Lipson.
\newblock How transferable are features in deep neural networks?
\newblock In Z.~Ghahramani, M.~Welling, C.~Cortes, N.~D. Lawrence, and K.~Q.
  Weinberger, editors, {\em Advances in Neural Information Processing Systems
  27}, pages 3320--3328. Curran Associates, Inc., 2014.

\bibitem[\protect\citeauthoryear{Zhang \bgroup \em et al.\egroup
  }{2017}]{DBLP:journals/corr/abs-1710-10577}
Quanshi Zhang, Wenguan Wang, and Song{-}Chun Zhu.
\newblock Examining {CNN} representations with respect to dataset bias.
\newblock {\em CoRR}, abs/1710.10577, 2017.

\bibitem[\protect\citeauthoryear{Zhang \bgroup \em et al.\egroup
  }{2018}]{Zhang_2018_CVPR}
Quanshi Zhang, Ying Nian~Wu, and Song-Chun Zhu.
\newblock Interpretable convolutional neural networks.
\newblock In {\em The IEEE Conference on Computer Vision and Pattern
  Recognition (CVPR)}, June 2018.

\bibitem[\protect\citeauthoryear{Zhang \bgroup \em et al.\egroup
  }{2020}]{zhang2020interpretable}
Quanshi Zhang, Xin Wang, Ying~Nian Wu, Huilin Zhou, and Song-Chun Zhu.
\newblock Interpretable cnns for object classification, 2020.

\bibitem[\protect\citeauthoryear{Zhou \bgroup \em et al.\egroup
  }{2014}]{DBLP:journals/corr/ZhouKLOT14}
Bolei Zhou, Aditya Khosla, {\`{A}}gata Lapedriza, Aude Oliva, and Antonio
  Torralba.
\newblock Object detectors emerge in deep scene cnns.
\newblock {\em CoRR}, abs/1412.6856, 2014.

\bibitem[\protect\citeauthoryear{Zhou \bgroup \em et al.\egroup
  }{2017}]{zhou2017places}
Bolei Zhou, Agata Lapedriza, Aditya Khosla, Aude Oliva, and Antonio Torralba.
\newblock Places: A 10 million image database for scene recognition.
\newblock {\em IEEE Transactions on Pattern Analysis and Machine Intelligence},
  2017.

\bibitem[\protect\citeauthoryear{Zhou \bgroup \em et al.\egroup
  }{2018}]{Zhou2018InterpretingDV}
Bolei Zhou, David Bau, Aude Oliva, and Antonio Torralba.
\newblock Interpreting deep visual representations via network dissection.
\newblock {\em IEEE transactions on pattern analysis and machine intelligence},
  2018.

\end{thebibliography}
	
\end{document}